%% file: neurips_2025.tex
\documentclass{article}

\PassOptionsToPackage{numbers, compress}{natbib}

\usepackage[preprint]{neurips_2025}

\usepackage[utf8]{inputenc} 
\usepackage[T1]{fontenc}    
\usepackage{hyperref}       
\usepackage{url}            
\usepackage{booktabs}       
\usepackage{amsfonts}       
\usepackage{nicefrac}       
\usepackage{microtype}      
\usepackage{xcolor}         
\usepackage{amsfonts}       
\usepackage{amsmath}        
\usepackage{enumitem}
\usepackage{graphicx}
\usepackage{multirow}

\title{Hallo4: High-Fidelity Dynamic Portrait Animation via Direct Preference Optimization}

\author{
Jiahao Cui$^{1\ast}$,\;\; 
Yan Chen$^{1\ast}$,\;\;
Mingwang Xu$^{1}$\thanks{indicates equal contribution.}\;,\;\;
Hanlin Shang$^{1}$,\;\;
Yuxuan Chen$^{1}$\\
\textbf{Yun Zhan}$^{1}$,\;\;
\textbf{Zilong Dong}$^{5}$,\;\;
\textbf{Yao Yao}$^{4}$,\;\;
\textbf{Jingdong Wang}$^{2}$,\;\;
\textbf{Siyu Zhu$^{1, 3}$}\\
$^{1}$Fudan University\;\;
$^{2}$Baidu Inc.\;\;
$^{3}$Shanghai Innovative Institute\;\;\\
$^{4}$Nanjing University\;\;
$^{5}$Alibaba Group
}

\begin{document}

\maketitle

\begin{abstract}
Generating highly dynamic and photorealistic portrait animations driven by audio and skeletal motion remains challenging due to the need for precise lip synchronization, natural facial expressions, and high-fidelity body motion dynamics.
We propose a human-preference-aligned diffusion framework that addresses these challenges through two key innovations.
First, we introduce direct preference optimization tailored for human-centric animation, leveraging a curated dataset of human preferences to align generated outputs with perceptual metrics for portrait motion-video alignment and naturalness of expression. 
Second, the proposed temporal motion modulation resolves spatiotemporal resolution mismatches by reshaping motion conditions into dimensionally aligned latent features through temporal channel redistribution and proportional feature expansion, preserving the fidelity of high-frequency motion details in diffusion-based synthesis.
The proposed mechanism is complementary to existing UNet and DiT-based portrait diffusion approaches, and experiments demonstrate obvious improvements in lip-audio synchronization, expression vividness, body motion coherence over baseline methods, alongside notable gains in human preference metrics. Our model and source code can be found at: https://github.com/fudan-generative-vision/hallo4.

\end{abstract}

\input{section/introduction}
\input{section/relatedwork}
\input{section/method}

\input{section/experiment}
\input{section/conclusion}
{\small
\bibliographystyle{plain}
\bibliography{egbib}
}

\end{document}

%% file: section/introduction.tex
\begin{figure*}
    \centering
    \includegraphics[width=0.97\linewidth]{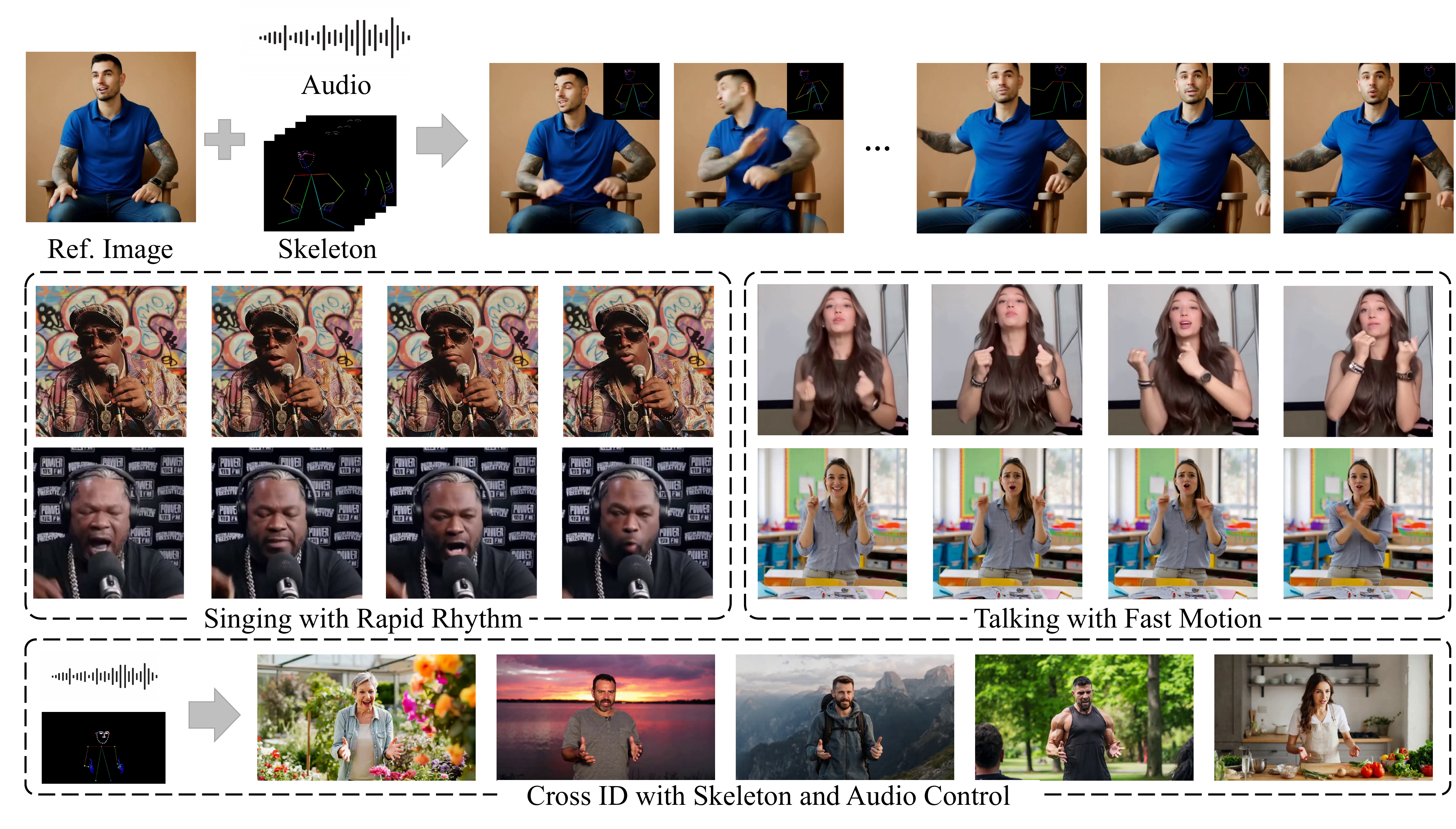}
    \caption{Illustration of the proposed portrait animation framework. Given a reference portrait image and multimodal control signals (audio waveform with optional skeletal motion sequences), our method generates high-fidelity, dynamically coherent animations through two key innovations: (1) direct preference optimization for human-aligned synchronization and expressiveness, and (2) unified temporal motion modulation to preserve high-frequency body motion details. 
    The framework achieves accurate lip-audio synchronization, natural facial expressions, and robust handling of rapid speech rhythms and abrupt upper-body motions across diverse character identities and environmental scenarios.}
    \label{fig:teaser}
\end{figure*}

\section{Introduction}
Portrait image animation, an interdisciplinary challenge at the intersection of computer vision and computer graphics, aims to synthesize photorealistic 2D videos of humans driven by multimodal control signals such as text, audio, and pose.
This technology holds significant potential for applications spanning digital entertainment, virtual reality, human-computer interaction, and personalized marketing. 
Despite recent advances in diffusion models~\cite{esser2024scaling,ho2020denoising,zhang2023adding}, vision transformers~\cite{liu2024vmamba,kirillov2023segment,liu2022swin}, and autoregressive architectures~\cite{touvron2023llama,alayrac2022flamingo,li2023blip}, achieving high-fidelity animation remains constrained by two persistent challenges: 
(1) generating lip-sync and facial expressions that are both perceptually natural and aligned with human preferences, 
and (2) capturing high-frequency motions, such as subtle articulations, dynamic expressions, and rapid body gestures—particularly during abrupt speech or sudden hand movements. 
In this work, we specifically investigate audio-driven animation of facial and upper-body portraits enhanced by skeletal motion conditions, aiming to address these two key challenges.

Direct preference optimization (DPO)~\cite{zhang2024direct}, a policy alignment framework that bypasses explicit reward modeling by optimizing generative models directly through pairwise human preference data, has gained increasing attention in text-to-image and text-to-video generation domains~\cite{wallace2024diffusion}. 
Unlike conventional reinforcement learning approaches, DPO minimizes distributional shifts during generation by aligning outputs with human preferences—such as aesthetic quality, semantic coherence, and temporal consistency—while eliminating the need for explicit reward model training and adversarial optimization loops through pairwise preference rankings with minimal data. 
Building on this paradigm, we introduce the first audio-driven portrait DPO dataset specifically designed for human-centric animation, capturing human preferences along two critical axes: 
(1) portrait-motion-video synchronization accuracy and (2) facial expression and pose naturalness. 
By contrasting trajectory probabilities of preferred and dispreferred samples using a KL-constrained reward function, our method optimizes the generation policy to maximize trajectory-level reward margins while regularizing deviations from the base diffusion model’s denoising dynamics. 
This approach achieves measurable improvements in lip-sync accuracy (matching real-data benchmarks in lip-sync scores) and enhances facial expressiveness, establishing a new baseline for preference-aligned portrait animation.

While DiT-based diffusion models have shown promise in outperforming UNet architectures in photorealistic portrait animation—particularly in generalizing across diverse scenes and character identities, achieving naturalistic character rendering, and seamlessly integrating portraits with environmental contexts—their dependence on temporally downsampled VAE-based features introduces a critical limitation.
Specifically, the standard practice of temporally downsampling motion conditions (e.g., lip articulation, facial expressions, and skeleton motions) to align with latent video temporal dimensions inevitably discards high-frequency and fine-grained motion details crucial for synchronizing rapid lip and gesture motions. 
To address these challenges, we propose a temporal motion condition modulation mechanism that unifies all motion conditions to match the temporal dimension of video latents while proportionally expanding feature dimensions to preserve motion granularity. 
This mechanism establishes a scalable foundation for integrating diverse motion signals with varying sampling frequencies into pretrained diffusion transformers, resolving fidelity degradation caused by latent-space temporal compression.

The proposed mechanism demonstrates complementary advantages when integrated with both UNet and DiT architectures. 
Our framework, combining direct preference optimization with unified temporal motion modulation, establishes a robust foundation for high-fidelity portrait animation. 
Comprehensive experiments demonstrate that our approach achieves significant improvements in lip-audio synchronization and expression naturalness compared to baseline methods, along with notable gains in human preference metrics, while maintaining superior motion coherence in complex upper-body gestures—particularly in challenging scenarios involving rapid speech, abrupt facial expressions and pose variations, and fast-changing body movements. 
To facilitate reproducible research in preference-aligned portrait generation, we will release both the codebase and curated preference dataset upon publication.

%% file: section/relatedwork.tex
\begin{figure*}
    \centering
    \includegraphics[width=.98\linewidth]{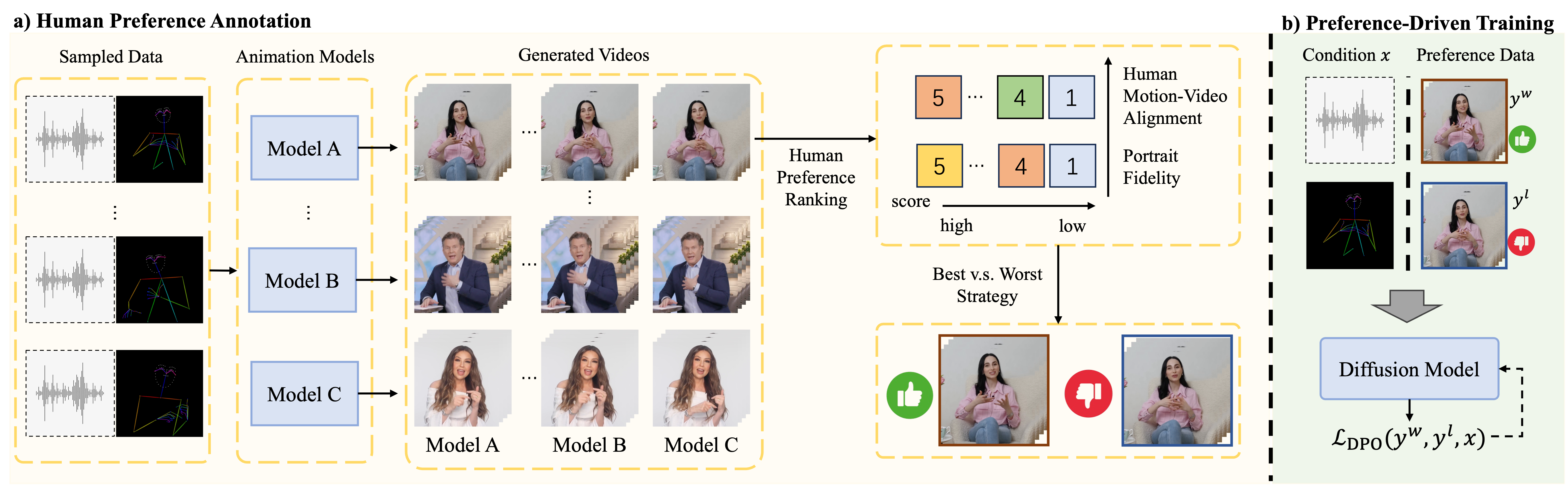}
    \caption{Demonstration of direct preference optimization for audio-driven portrait animation.}
    \label{fig:pipeline}
    \vspace{-4mm}
\end{figure*}

\section{Related Work}
\textbf{Portrait Image Animation.}
Research in portrait animation has evolved through distinct technical paradigms addressing sequential limitations. 
Early lip-synchronization methods, exemplified by Wav2Lip~\cite{prajwal2020lip}, employed adversarial discriminators to enhance audio-visual alignment but lacked holistic facial dynamics. 
Subsequent 3D morphable model (3DMM)-based approaches improved motion naturalness through structured representations: 
SadTalker~\cite{zhang2022sadtalker} disentangled audio-driven 3D coefficients, while VividTalk~\cite{sun2023vividtalk} and DreamTalk~\cite{zhang2023dream} introduced hybrid mesh-codebook strategies. 
Diffusion-driven frameworks advanced realism by integrating facial landmarks (DiffTalk~\cite{shen2023difftalk}, AniPortrait~\cite{wei2024aniportrait}) and hierarchical temporal conditioning (Loopy~\cite{jiang2024loopy}, Hallo~\cite{xu2024hallo}), with LivePortrait~\cite{guo2024liveportrait} optimizing efficiency via implicit keypoints. 
Audio-centric methods like Sonic~\cite{ji2024sonic} prioritized global perceptual coherence for diverse motion synthesis. 
Transformer-based architectures, including VASA-1~\cite{xu2024vasa}, Hallo3~\cite{cui2024hallo3} and OmniHuman~\cite{lin2025omnihuman}, unified facial and bodily dynamics in latent spaces for partial and full-body generalization.
Our work introduces a human-annotated DPO dataset and optimization approach to enhance lip-sync accuracy and facial naturalness, complementing existing UNet and DiT-based audio-driven diffusion frameworks.

\textbf{Human Preference Alignment.}
Research on aligning generative models with human preferences spans visual understanding and synthesis tasks. 
In video understanding, early efforts~\cite{sun2023aligning} adapted reinforcement learning with human feedback to mitigate hallucinations, while subsequent works introduced self-supervised alignment via language model rewards~\cite{zhang2024direct}, AI feedback~\cite{ahn2024tuning}, and temporal grounding optimization~\cite{li2025temporal}. 
For image generation, Diffusion-DPO~\cite{wallace2024diffusion} reformulated DPO to align text-to-image outputs with human judgments. 
Video generation methods further extended DPO to optimize visual-textual semantics (VideoDPO~\cite{liu2024videodpo}), enable online training (OnlineVPO~\cite{zhang2024onlinevpo}), and refine motion dynamics (Flow-DPO~\cite{liu2025improving}). 
While these works establish DPO as a scalable paradigm for preference-aware synthesis, none address the unique challenges of audio-driven portrait animation. 
To our knowledge, this work presents the first curated human preference dataset and tailored DPO framework for enhancing lip-sync accuracy and facial naturalness in portrait animation domain.

\textbf{Motion Condition.} 
Strategies for motion condition injection in human video generation depend on motion signal type and control granularity. 
Audio-driven portrait methods, such as EMO~\cite{tian2021good}, VASA-1~\cite{xu2024vasa}, and FantasyTalking~\cite{wang2025fantasytalking}, leverage cross-attention to align audio features with visual frames. 
Body motion control approaches like CHAMP~\cite{zhu2024champ}, MimicMotion~\cite{zhang2024mimicmotion}, and EasyControl~\cite{wang2024easycontrol} employ in-context learning for coarse-grained skeletal dynamics. 
Half-body animation frameworks, including CyberHost~\cite{lin2024cyberhost}, EchoMimicV2~\cite{meng2024echomimicv2}, and OmniHuman~\cite{lin2025omnihuman}, combine cross-attention for facial synchronization with in-context learning for skeletal motion, achieving coherent upper-body motion. 
Building on these insights, we propose a temporal motion condition modulation mechanism that enhances high-frequency motion details and fine-grained articulatory synchronization, particularly optimizing motion fidelity for DiT-based portrait diffusion models.

%% file: section/method.tex
\begin{figure*}
    \centering
    \includegraphics[width=.98\linewidth]{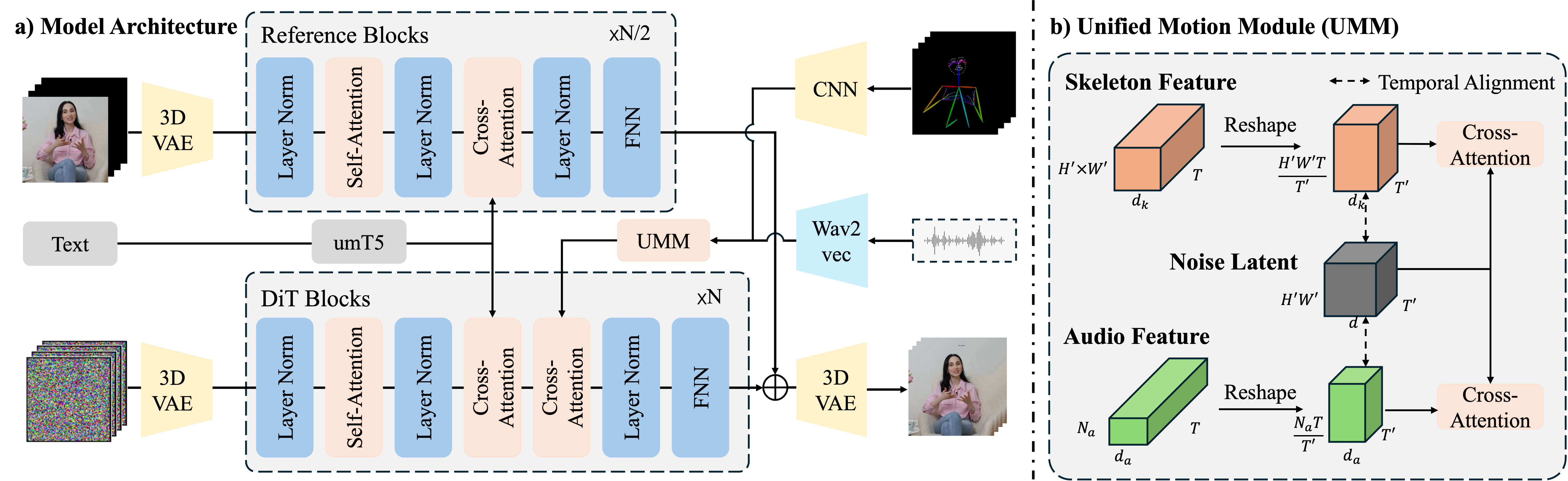}
    \caption{Demonstration of a DiT-based portrait generative pipeline with unified temporal motion modulation.}
    \label{fig:pipeline}
\end{figure*}

\section{Methodology}
Our framework integrates human preference driven diffusion optimization and unified temporal motion modulation for highly dynamic and high-fidelity portrait animation.
Section~\ref{sec:preliminary} presents the preliminary of diffusion models.
Section~\ref{sec:animation} introduces curated preference datasets that capture motion-video alignment and portrait fidelity through best-vs-worst ranking and adapts direct preference optimization to align denoising trajectories with human judgments. 
Section~\ref{sec:motion} resolves temporal resolution mismatches via unified motion modulation through feature reshaping and cross-attention fusion. 
Finally, Section~\ref{sec:network} implements this through diffusion networks with DiT/UNet backbones, enabling coherent synthesis of audio-driven facial movements and skeleton-guided body dynamics.

\subsection{Preliminary}\label{sec:preliminary}
Diffusion models generate data through an iterative denoising process governed by a forward noise-addition process and a learned reverse process. Formally, given a data sample $\mathbf{x}_0 \sim q(\mathbf{x}_0)$, the forward process gradually perturbs the data with Gaussian noise over $T$ timesteps according to:
\begin{equation}
    q(\mathbf{x}_t|\mathbf{x}_{t-1}) = \mathcal{N}(\mathbf{x}_t; \sqrt{1-\beta_t}\mathbf{x}_{t-1}, \beta_t\mathbf{I}),
\end{equation}
where $\{\beta_t\}_{t=1}^T$ defines a predetermined noise schedule. This yields the marginal distribution at timestep $t$:
\begin{equation}
    \mathbf{x}_t = \sqrt{\bar{\alpha}_t}\mathbf{x}_0 + \sqrt{1-\bar{\alpha}_t}\boldsymbol{\epsilon},
\end{equation}
with $\bar{\alpha}_t = \prod_{s=1}^t(1-\beta_s)$ and $\boldsymbol{\epsilon} \sim \mathcal{N}(\mathbf{0}, \mathbf{I})$. The reverse process learns to denoise through a neural network $\epsilon_\theta$ parameterized by $\theta$, optimized to predict the noise component via:
\begin{equation}
    \mathcal{L}(\theta) = \mathbb{E}_{t,\mathbf{x}_0,\boldsymbol{\epsilon}}\left[\|\boldsymbol{\epsilon} - \epsilon_\theta(\mathbf{x}_t,t)\|^2\right].
\end{equation}
Conventional implementations employ U-Net architectures for $\epsilon_\theta$, leveraging their encoder-decoder structure with skip connections. 
Recent advances propose diffusion transformers as competitive alternatives, utilizing self-attention mechanisms to enhance global coherence—a critical feature for modeling video frame dependencies.  

Our framework builds upon pretrained portrait diffusion models, augmented by direct preference optimization to align animations with human preferences. 
Another pivotal challenge lies in effectively integrating motion conditions with varying temporal frequencies into the noise latent space, which is temporally compressed via a VAE. 
This integration is essential for achieving highly dynamic and naturalistic portrait animations.

\subsection{Preference-Driven Portrait Animation}\label{sec:animation}
To establish a robust foundation for direct preference optimization in portrait animation, we curate a multimodal dataset that captures human preferences across two critical dimensions: (1) motion-video alignment and (2) portrait fidelity.

\textbf{Preference Metrics.} 
Motion-video alignment and portrait fidelity constitute two critical metrics for evaluating synthesized portrait animations. 
Motion-video alignment quantifies the temporal consistency between generated animations and human motion conditions, emphasizing lip synchronization with audio input and body movements synchronized to rhythmic beats. 
Portrait fidelity evaluates the perceptual quality of synthesized content, with particular emphasis on the semantic coherence of facial expressions and head poses relative to speech content (e.g., conveying emotions such as joy, anger, or surprise), as well as the overall naturalness and realism of the rendered output. 
These metrics collectively ensure both perceptual accuracy and human-centric plausibility in dynamic portrait generation systems.

\textbf{Preference Dataset Curation.}
Direct preference optimization frames human preference alignment as a policy optimization problem using pairwise preference data $\mathcal{D} = \{(x, y^w, y^l)\}$, where $y^w$ is preferred over $y^l$; here, $y^w$ and $y^l$ represent the ``winning'' and ``losing'' samples, respectively~\cite{wallace2024diffusion}. 
To construct this dataset, we employ annotators to assess the generated samples across the two dimensions and combine them with a ``best-vs-worst'' selection strategy to obtain the final preference dataset. 

For each input audio clip, we generate candidate videos using five representative methods spanning GAN-based (SadTalker), UNet-based diffusion (AniPortrait, EchoMimic-v2), and DiT-based diffusion (FantasyTalking, Hallo3) architectures. 
Following independent generation human annotators evaluate each video on two critical axes: motion-video alignment (lip/body synchronization) and portrait fidelity (expression naturalness, rendering quality), assigning scores on a 5-point Likert scale. 
To aggregate the scores across both dimensions, a composite reward is defined as  $r = \frac{1}{2}(r_{\text{align}} + r_{\text{fidelity}})$. 

Pairwise preferences data $(y^w, y^l)$ are derived by ranking samples based on composite rewards $r$ with ``best-vs-worst'' strategy. We define the sample with the highest reward as $y^w$, and the one with the lowest as $y^l$. 
This strategy maximizes the reward margin between $y^w$ and $y^l$ samples, encouraging clear distinctions in subtle aspects like lip sync and expression quality. 
To maintain the discriminability of the dataset, cases with minimal reward differences are excluded. 

\textbf{Preference Optimization.}
To align generated animations with human preferences, we adapt direct preference optimization to diffusion-based portrait synthesis. 
Let the generation policy $\pi_\theta(y|x)$ represents the conditional distribution of videos $y$ given input conditions $x$ (audio, skeleton), and $\pi_{\text{ref}}$ denotes a reference policy initialized from a supervised fine-tuned diffusion model. 
The DPO objective maximizes the likelihood of preferred outputs while regularizing deviations from $\pi_{\text{ref}}$:
\begin{equation}
    \mathcal{L}_{\text{DPO}} = -\mathbb{E}_{(x,y^w,y^l)\sim\mathcal{D}} \left[ \log\sigma\left( \beta \log\frac{\pi_\theta(y^w|x)}{\pi_{\text{ref}}(y^w|x)} - \beta \log\frac{\pi_\theta(y^l|x)}{\pi_{\text{ref}}(y^l|x)} \right) \right],
\end{equation}  
where $\sigma(\cdot)$ is the sigmoid function and $\beta$ controls deviation from $\pi_{\text{ref}}$. 
This objective implicitly defines the reward function $r_\theta(x,y) = \beta \log\frac{\pi_\theta(y|x)}{\pi_{\text{ref}}(y|x)}$, circumventing explicit reward modeling by leveraging pairwise preference datasets.

For DiT diffusion models via flow-matching, the DPO loss is formulated as:
\begin{equation}
\begin{aligned}
&\mathcal{L}_{\text{DPO}} = -\mathbb{E}_{t \sim \mathcal{U}(0, 1),\ (y^w, y^l) \sim \mathcal{D}} \log \sigma\Big( 
    -\beta \omega_t \big[ \|v^w - v_{\theta}(y_t^w, t)\|^2 \\
    &- \|v^w - v_{\text{ref}}(y_t^w, t)\|^2 
    - \big( \|v^l - v_{\theta}(y_t^l, t)\|^2 
    - \|v^l - v_{\text{ref}}(y_t^l, t)\|^2 \big) \big] \Big)
\end{aligned}
\end{equation}
where $\omega_t$ is a weighting function, $v^w$ and $v^l$ denote the velocity fields derived from preferred sample $y^w$ and dispreferred sample $y^l$, respectively. 
The expectation is taken over the preference dataset $\mathcal{D} = \{(x, y^w, y^l)\}$ and noise timestep $t$. 
By minimizing $\mathcal{L}_{\text{DPO}}$, the predicted velocity field $v_{\theta}$ becomes closer to $v^w$ (aligned with human preferences) and further from $v^l$, promoting coherent motion trajectories and produce more natural and realistic expressions.

During training, gradients from $\mathcal{L}_{\text{DPO}}$ adjust the denoising direction to favor high-reward regions in the trajectory space while preserving the temporal dynamics of the reference policy. 
This joint optimization enables preference-aware generation without compromising the inherent stability of pretrained diffusion models.

\subsection{Unified Temporal Motion Modulation}\label{sec:motion}
Existing DiT-based diffusion models face challenges in aligning multimodal motion signals (e.g., speech audio and skeletal sequences) with compressed video latent representations, primarily due to temporal resolution mismatches and dimensional incompatibility. 
To address this, we propose a temporal reshaping strategy that aligns motion conditions with video latents through temporal dimension matching and proportional feature channel expansion, effectively preserving motion granularity while maintaining dimensional consistency.

\textbf{Latent Video Representation.}
The original video data $\mathbf{V}$ is encoded into a compressed latent representation $\mathbf{Z}$ via a pretrained causal 3D variational autoencoder. 
Following established methods~\cite{wan2025}, the input video $\mathbf{V} \in \mathbb{R}^{T \times H \times W \times 3}$ is compressed to latent features $\mathbf{Z} = \mathcal{E}(\mathbf{V}) \in \mathbb{R}^{T' \times H' \times W' \times d}$, where $T' = \lfloor T/4 \rfloor$ represents the temporally downsampled length, and $H' = \lfloor H/8 \rfloor$, $W' = \lfloor W/8 \rfloor$ denote spatially reduced dimensions. 
This encoding balances computational efficiency with the preservation of critical motion dynamics and spatiotemporal coherence inherent in the original video.

\textbf{Human Motion Condition.}
We process two complementary motion modalities to govern facial articulations and body dynamics. 
For audio-driven facial movements, the raw speech signals $\mathbf{S} \in \mathbb{R}^L$ are encoded into frame-aligned acoustic features $\mathbf{C}_{\text{audio}} \in \mathbb{R}^{T \times N_a \times d_a}$ via Wav2Vec 2.0~\cite{baevski2020wav2vec}, preserving phoneme-level temporal precision essential for lip synchronization. 
For skeletal-guided body motion, 2D human keypoints $\mathbf{V}_k \in \mathbb{R}^{T \times J \times 3}$ (with $J$ joints) are processed through an efficient convolutional network, generating motion descriptors $\mathbf{C}_{\text{skel}} \in \mathbb{R}^{T \times H' \times W' \times d_k}$ that maintain original temporal resolution while achieving spatial alignment with video latents through dimension reduction to $H' \times W'$. 
This dual-conditioning mechanism encourages synchronized integration of high-frequency facial movements and coarse body kinematics with the latent video representation.

\textbf{Temporal Reshaping and Fusion.}
To resolve temporal resolution mismatches between motion conditions (frame rate $T$) and compressed video latents (downsampled length $T'$), we propose a feature redistribution strategy that preserves motion fidelity through dimensional expansion. 
For each motion condition $\mathbf{C}_m \in \mathbb{R}^{T \times D_m \times d_m}$ (where $m \in \{\text{audio}, \text{skel}\}$), we reshape the temporal axis via:
\begin{equation}
    \tilde{\mathbf{C}}_m = \text{Reshape}\left(\mathbf{C}_m, \left(T', \rho D_m\right)\right) \in \mathbb{R}^{T' \times (\rho D_m) \times d_m},
\end{equation}
where $\rho = T/T'$ denotes the latent compression ratio. 
This operation eliminates subsampling artifacts by distributing original temporal features into expanded channel dimensions while maintaining alignment with latent length $T'$. 
The reshaped features undergo latent-space projection through learnable linear transformations:
\begin{equation}
    \mathbf{H}_m = \mathbf{W}_m \tilde{\mathbf{C}}_m + \mathbf{b}_m \in \mathbb{R}^{T' \times (\rho D_m) \times d},
\end{equation}
ensuring dimensional compatibility with the video latents $\mathbf{Z} \in \mathbb{R}^{T' \times H' \times W' \times d}$. 
Multimodal fusion is achieved via cross-attention:
\begin{equation}
    \mathbf{Z}' = \text{Softmax}\left(\frac{\mathbf{Q}\mathbf{K}^\top}{\sqrt{d}}\right)\mathbf{V} + \mathbf{Z},
\end{equation}
where $\mathbf{Q} = \mathbf{Z}\mathbf{W}_Q$, $\mathbf{K} = \mathbf{H}\mathbf{W}_K$, $\mathbf{V} = \mathbf{H}\mathbf{W}_V$, and $\mathbf{H} = [\mathbf{H}_{\text{audio}} \| \mathbf{H}_{\text{skel}}]$. 
This design offers dual advantages: 
1) High-frequency motion details critical for lip synchronization and subtle gestures are preserved through feature expansion rather than temporal subsampling; 
2) Dimensionally consistent projections enable seamless integration with pretrained diffusion transformers without architectural modifications.

\subsection{DPO-Enhanced Diffusion Network}\label{sec:network}

\textbf{Architecture.}
As illustrated in Figure~\ref{fig:pipeline}, we validate our preference dataset and DPO framework on both UNet-based and DiT-based diffusion architectures. 
The proposed unified temporal motion modulation strategy is specifically implemented within the DiT-based diffusion framework, where temporal compression in the noise latent space is achieved via a variational autoencoder.

For the UNet architecture, we adopt the design principles of EMO~\cite{tian2024emo}, Hallo~\cite{xu2024hallo}, and EchoMimic-v2~\cite{chen2024echomimic}, using Stable Diffusion 1.5~\cite{blattmann2023stable} as the backbone. 
A parallel ReferenceNet structure preserves visual identity consistency, while audio-conditioning layers inject speech features through cross-attention mechanisms. 
The diffusion denoising process employs denoising diffusion probabilistic models, which learns a parameterized reverse process through variational lower bound optimization to iteratively recover clean data from noise.

For the DiT architecture, we utilize the Wan2.1 framework~\cite{wan2025} as the backbone, consistent with FantasyTalking~\cite{wang2025fantasytalking}. 
This architecture employs a 3D causal VAE for spatiotemporal compression and decoding. 
The transformer module processes video data by unfolding it into token sequences via 3D patchify operations. 
Each transformer block integrates self-attention for spatial relations, cross-attention for text conditioning, and timestep embeddings. 
All transformer layers share a unified MLP structure. 
The denoising process leverages flow matching principles, constructing vector fields based on optimal transport path conditional probability flows to directly regress velocity fields. 
This enables deterministic ODE-based transformations from noise distributions to data distributions.

\subsection{Training and Inference}

\textbf{Training.}
The training protocol adapts to architectural distinctions between UNet and DiT backbones while maintaining preference optimization consistency. 

For UNet-based diffusion, we initialize with the pretrained Hallo portrait model~\cite{xu2024hallo} and fine-tune it using direct preference optimization. 
The DPO objective $\mathcal{L}_{\text{DPO}}$ optimizes denoising trajectories to align with human preference judgments through gradient updates on the UNet parameters.

For DiT architectures, training follows a phased approach to ensure stable integration of multimodal conditions. 
Phase 1 focuses on audio-driven synthesis, where parameter updates are restricted to the cross-attention layers in the unified motion module, keeping the 3D VAE and denoiser blocks frozen. 
Phase 2 introduces skeletal guidance by extending the motion module to process joint kinematics, updating only the newly introduced skeletal cross-attention parameters while maintaining frozen 3D VAE and denoiser block weights. 
Both phases leverage flow matching to regress velocity fields encoding spatiotemporal dynamics. 
Finally, we apply DPO training that adjusts the denoising policy through comparative gradient updates, prioritizing human-preferred motion patterns and visual qualities while preserving the model's temporal coherence.

\textbf{Inference.}
During inference, the model integrates four input modalities: a reference image for visual context, a driving audio segment for speech signals, a textual prompt for semantic guidance, and skeleton frames encoding body motion dynamics. 
It generates a temporally coherent portrait animation by harmonizing audio-driven lip synchronization with skeleton-guided pose dynamics, encouraging both expressive facial movements and anatomically consistent body kinematics.

%% file: section/experiment.tex
\begin{figure}
    \centering
    \includegraphics[width=0.98\linewidth]{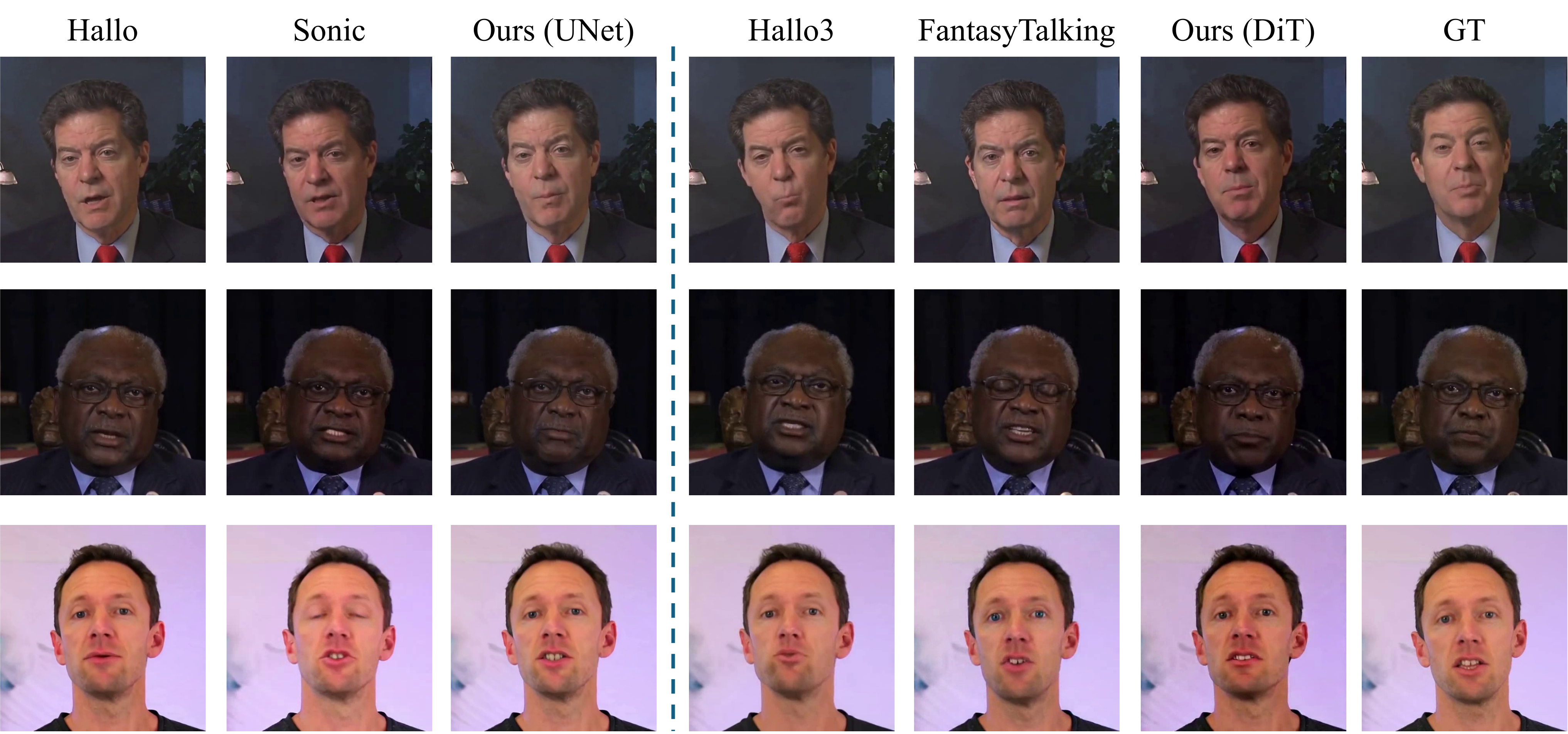}
    \caption{Qualitative comparison on HDTF and Celeb-V dataset.}
    \label{fig:comp-hdtf-celebv}
\end{figure}

\section{Experiment}
\subsection{Experimental Setups}
\textbf{Implementation Details.}
Our framework was implemented on 8 NVIDIA A100 GPUs through two sequential training phases. 
The temporal motion modulation module underwent initial training with 10,000 iterations using a learning rate of 1e-5, optimized via AdamW with 2,000 warm-up steps to stabilize gradient updates. 
A batch size of 8 ensured memory efficiency while preserving motion dynamics. For direct preference optimization, we trained for 12,000 iterations with a reduced learning rate of 1e-8 and 2,500 warm-up steps, employing a beta parameter of 2500 to balance preference alignment. 
The DPO phase maintained a global batch size of 8.

\textbf{Datasets.}
Our training dataset comprises 220K source videos (220 hours) sourced from diverse platforms including Celeb-V, HDTF, and YouTube, encouraging coverage across varied demographic and environmental conditions. 
From this collection, we curate 20K high-confidence video pairs through a rigorous selection process, where samples with the minimal composite reward score are excluded to emphasize quality differentials in portrait animation. 
The source videos train the base DiT portrait animation model, while the curated pairs facilitate direct preference optimization. 
For evaluation, we employ three benchmarks: HDTF for general talking-head synthesis, CelebV-HQ for high-resolution portrait generation, and EMTD to assess motion diversity and anatomical consistency in skeletal-driven animations.

\begin{table}[th!]
    \centering
    \begin{tabular}{c|c|c|c|c}
    \toprule
    \textbf{Architecture} & \textbf{Method} & \textbf{Sync-C$\uparrow$} & \textbf{Sync-D$\downarrow$} & \textbf{E-FID$\downarrow$} \\
    \midrule
    \multirow{6}{*}{UNet-based}
    &SadTalker~\cite{zhang2022sadtalker}          & 7.804  & 7.956 & 11.826 \\
    &DreamTalk~\cite{ma2023dreamtalk}             & 7.515 & 7.717 & 9.142 \\
    &AniPortrait~\cite{wei2024aniportrait}        & 3.550 & 11.007 & 14.819 \\
    &Sonic~\cite{ji2024sonic}                     & 8.186 & \textbf{6.823} & 9.370 \\
    &Hallo~\cite{xu2024hallo}                     & 7.770 & 7.605 & 8.508 \\
    &Ours~(UNet) & \textbf{8.286} & 7.502 & \textbf{8.241} \\
    \midrule
    \multirow{3}{*}{DiT-based}
    &Hallo3~\cite{cui2024hallo3}                  & 7.384 & 8.613 & 8.589 \\
    &FantasyTalking~\cite{wang2025fantasytalking} & 4.218 & 11.043 & 9.806 \\
    &Ours~(DiT)                                  & \textbf{9.161} & \textbf{6.987} & \textbf{7.645} \\
    \midrule
    \multicolumn{2}{c|}{Real Video}             & 8.976 & 6.359 & -- \\ 
    \bottomrule
    \end{tabular}
    \vspace{2mm}
    \caption{Comparison with existing audio-driven portrait animation methods on HDTF dataset. 
    For the UNet-based diffusion models, ``Ours (UNet)'' achieves superior motion-video alignment metrics in terms of lip motion synchronization score (Sync-C, Sync-D), compared to its baseline Hallo.
    For the DiT-based diffusion models, the performance improvement from DPO and unified temporal motion modulation is more obvious.}
    \label{tab:comp_hdtf}
\end{table}

\begin{table}[ht!]
    \centering
    \begin{tabular}{c|c|c|c|c|c|c}
    \toprule
    \textbf{Architecture} & \textbf{Method} & \textbf{Sync-C$\uparrow$} & \textbf{Sync-D$\downarrow$} & \textbf{E-FID$\downarrow$} & \textbf{FID$\downarrow$} & \textbf{FVD$\downarrow$} \\
    \midrule
    
    \multirow{6}{*}{UNet-based}
    & SadTalker~\cite{zhang2022sadtalker}          & 5.126  & 8.219 & 43.499 & 64.919 & 463.671 \\
    & DreamTalk~\cite{ma2023dreamtalk}             & 4.691 & 8.290 & 68.211 & 137.049 & 879.902 \\
    & AniPortrait~\cite{wei2024aniportrait}        & 2.129 & 10.911  & 31.760 & 63.106 & 509.498 \\
    & Sonic~\cite{ji2024sonic}                     & 5.387 & \textbf{7.370} & 19.928 & 56.104 & 489.458 \\
    & Hallo~\cite{xu2024hallo}                     & 4.880 & 8.189 & 27.252 & 51.009 & 469.985 \\
    & Ours~(Unet) & \textbf{5.480} & 8.389 & \textbf{19.627} & \textbf{49.282} & \textbf{444.319} \\ \midrule
    
    \multirow{3}{*}{DiT-based}
    & Hallo3~\cite{cui2024hallo3}               & 4.648 & 8.992 & 23.671 & \textbf{54.105} & \textbf{397.425} \\
    & FantasyTalking~\cite{wang2025fantasytalking} & 2.957 & 10.559 & 47.124 & 69.332 & 614.462 \\
    & Ours~(DiT)                                 & \textbf{5.689} & \textbf{7.853} & \textbf{18.998} & 58.815 & 538.396 \\
    \midrule
    
    \multicolumn{2}{c|}{Real Video}             & 4.550 & 8.090 & --      & --     & --      \\
    \bottomrule
    \end{tabular}
    \vspace{2mm}
    \caption{Comparison with existing audio-driven portrait animation methods on Celeb-V dataset.
    Given that the Celeb-V data are more wild than HDTF dataset, the performance improvement in terms of motion-video alignment and expression fidelity from DPO and and unified temporal motion modulation is more obvious than the corresponding UNet and DiT-based baselines.}
    \label{tab::celebv}
\end{table}

\subsection{Experimental Setups}

\textbf{Evaluation Metrics.}
For motion-video alignment, we adopt the widely used Sync-C and Sync-D metrics~\cite{chung2017out} to measure the confidence and distance between visual and audio content. 
Hand quality and motion richness are measured using hand keypoint confidence (HKC)~\cite{lin2024cyberhost} and hand keypoint variance (HKV)~\cite{lin2024cyberhost}, respectively.
These metrics validate the DPO results after integrating preference data with human motion alignment, as well as the improvements in highly dynamic motion enabled by the unified temporal motion modulation module. 
To evaluate facial fidelity, we employ E-FID~\cite{deng2019accurate}, which quantifies expression discrepancies by computing the Fréchet Inception Distance (FID) of facial expression parameters extracted via face reconstruction. 
For visual quality assessment, we utilize FID, FVD~\cite{unterthiner2018towards}, PSNR, and SSIM.

\textbf{Baseline.}
We conduct comprehensive comparisons across two dominant architectural paradigms for audio-driven portrait animation: UNet-based and DiT-based approaches. 
For UNet-based methods, we benchmark against SadTalker~\cite{zhang2022sadtalker}, DreamTalk~\cite{ma2023dreamtalk}, AniPortrait~\cite{wei2024aniportrait}, Sonic~\cite{ji2024sonic}, and Hallo~\cite{xu2024hallo}. 
Our UNet variant, denoted as Ours (UNet), extends Hallo by integrating human preference data and direct preference optimization (DPO). 
Within the DiT-based category, we evaluate Hallo3~\cite{cui2024hallo3} and FantasyTalking~\cite{wang2025fantasytalking}, while our DiT implementation (Ours (DiT)) builds upon the Wan2.1 backbone~\cite{wan2025}, incorporating unified temporal motion modulation and DPO-enhanced training.
To validate generalization to skeletal-guided synthesis, we further compare with pose-driven half-body animation methods: AnimateAnyone~\cite{hu2024animate}, MimicMotion~\cite{zhang2024mimicmotion}, and EchoMimic-v2~\cite{meng2024echomimicv2}. 

\subsection{Comparison with State-of-the-Art}
\textbf{Comparison on HDTF Dataset.}  
As shown in Table~\ref{tab:comp_hdtf}, our method achieves state-of-the-art performance on HDTF, outperforming existing approaches in critical metrics. 
Notably, the DiT variant achieves a Sync-C score of 9.161 and Sync-D of 6.987, surpassing even real videos (8.976 and 6.359), which underscores the efficacy of direct preference optimization in refining motion-video alignment and expression fidelity, as evidenced by the lowest E-FID (7.645).
More visual comparison is shown in Figure~\ref{fig:comp-hdtf-celebv}.

\textbf{Comparison on Celeb-V Dataset.}
On the more challenging Celeb-V dataset (Table~\ref{tab::celebv}), our method exhibits robust generalization, achieving superior Sync-C (5.689) and E-FID (18.998) compared to all baselines. 
The obvious Sync-C improvement (22.4\% over Hallo3) highlights DPO’s capability to adapt to wild scenarios by prioritizing human preferences for lip coherence and expression fidelity.

\begin{table}[t!]
    \centering
    \label{tab:comp_halfbody}
    \begin{tabular}{c|c|c|c|c|c|c|c}
    \toprule
     & \textbf{Sync-C$\uparrow$} & \textbf{HKV$\uparrow$} & \textbf{HKC$\uparrow$} & \textbf{PSNR$\uparrow$} & \textbf{SSIM$\uparrow$} & \textbf{FID$\downarrow$} & \textbf{FVD$\downarrow$} \\
    \midrule 
    AnimateAnyone~\cite{hu2024animate} & - & 34.350 & 0.824 & 20.482 & 0.761 & 51.439 & 453.528 \\
    MimicMotion~\cite{zhang2024mimicmotion} & - & 34.783 & 0.846 & 19.186 & 0.713 & 49.039 & 468.004 \\
    EchoMimic-v2~\cite{meng2024echomimicv2} & 6.624 & 33.284 & 0.863 & 21.421 & 0.796 & \textbf{38.461} & \textbf{301.347} \\
    \midrule
    Ours~(DiT) & \textbf{6.764} & \textbf{35.736} & \textbf{0.886} & \textbf{22.221} & \textbf{0.808} & 44.422 & 381.752 \\ 
    \bottomrule 
    \end{tabular}
    \vspace{2mm}
    \caption{Quantitative comparison of pose-driven half-body animation methods on EMTD dataset. 
    Our approach demonstrates superior lip synchronization (Sync-C) through direct preference optimization, while the unified motion modulation mechanism enhances hand articulation quality (HKC) and motion diversity (HKV).}
    \label{tab:comp_halfbody}   
\end{table}

\begin{figure}
    \centering
    \includegraphics[width=0.98\linewidth]{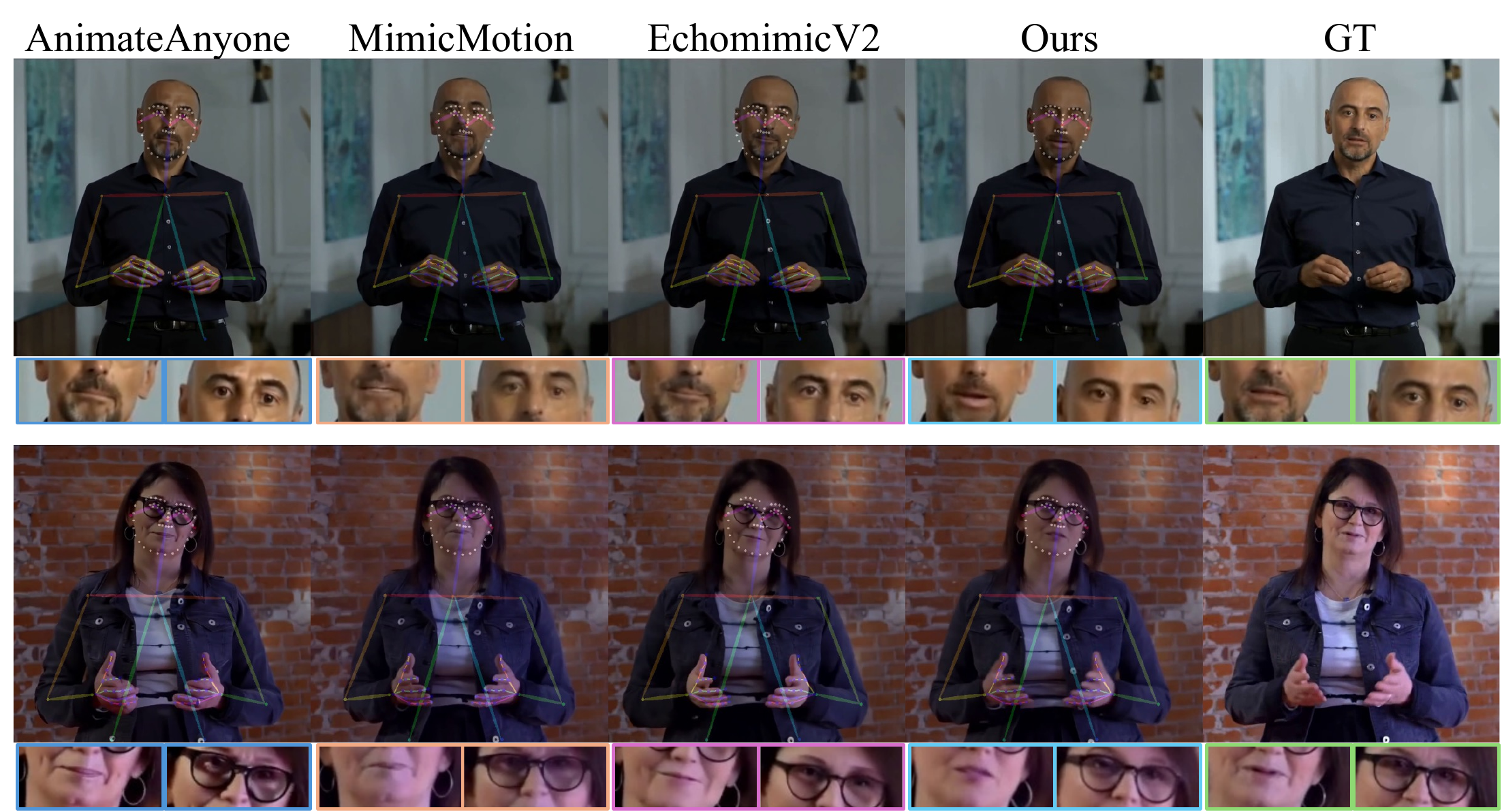} 
    \vspace{2mm}
    \caption{Qualitative comparison on half-body EMTD dataset.}
    \label{fig:comp-emtd}
\end{figure}

\textbf{Comparison on Half-Body EMTD Dataset.}
Our evaluation on the EMTD dataset for skeletal and audio-driven animations demonstrates that the unified temporal motion modulation obvious enhances hand motion diversity (HKV) and articulation quality (HKC), while improving visual fidelity (PSNR/SSIM). 
As shown in Table~\ref{tab:comp_halfbody}, our method outperforms pose-driven baselines by effectively aligning high-frequency skeletal dynamics with latent video representations, achieving a 35.736 HKV and 0.886 HKC, which underscores its capacity to harmonize anatomical consistency with motion granularity.
The qualitative comparison is shown in Figure~\ref{fig:comp-emtd}.

\begin{figure}
    \centering
    \includegraphics[width=1\linewidth]{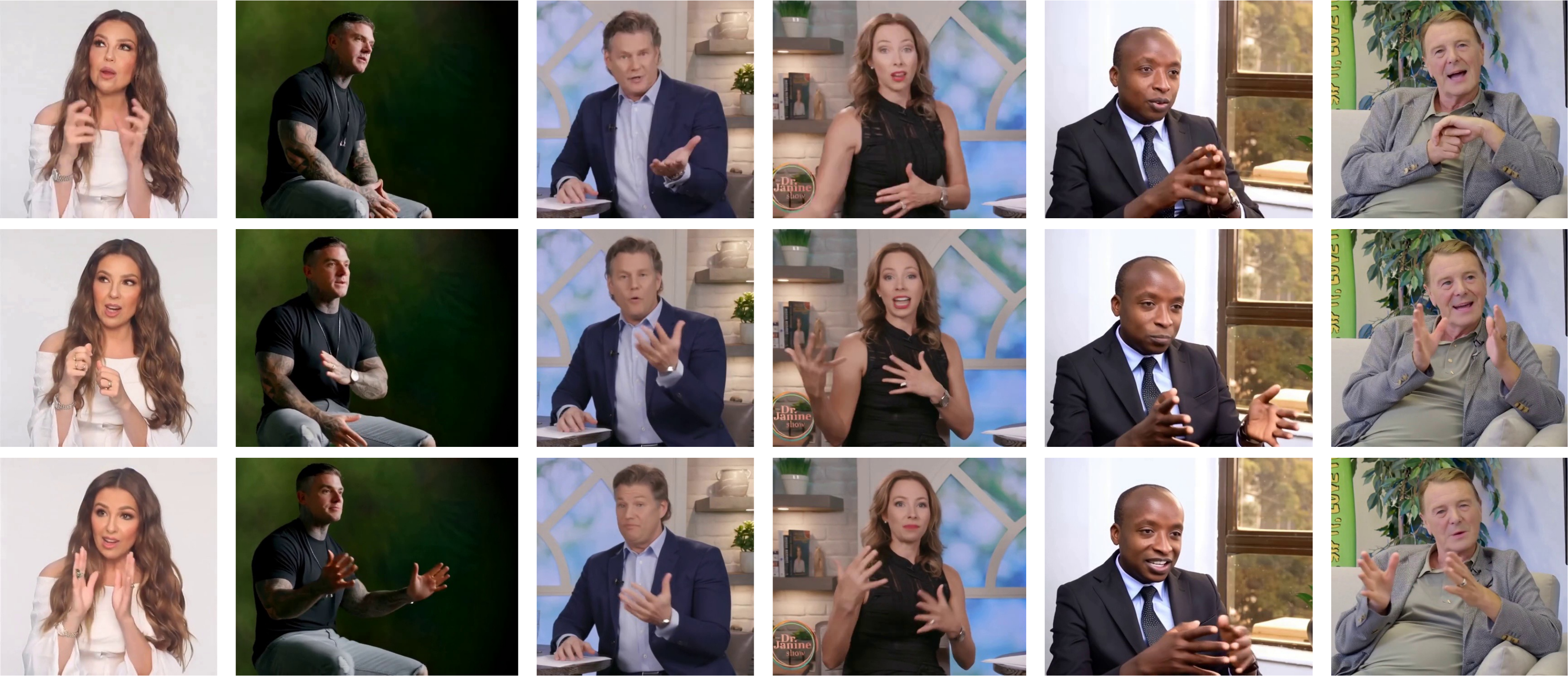}
    \vspace{2mm}
    \caption{Demonstration of more results from wild data-set.}
    \label{fig:wild-results}
\end{figure}

\textbf{Visualization on Wild Data-set.}
Figure~\ref{fig:wild-results} showcases our method’s generalization across diverse identities, environments, and motion intensities.
The generated animations maintain lip-audio synchronization and anatomical consistency even under challenging scenarios involving rapid speech rhythms (e.g., rap vocals), exaggerated body movements (e.g., dance sequences), and variable lighting conditions.

\subsection{Ablation Study and Discussion}

\begin{figure}
    \centering
    \includegraphics[width=1\linewidth]{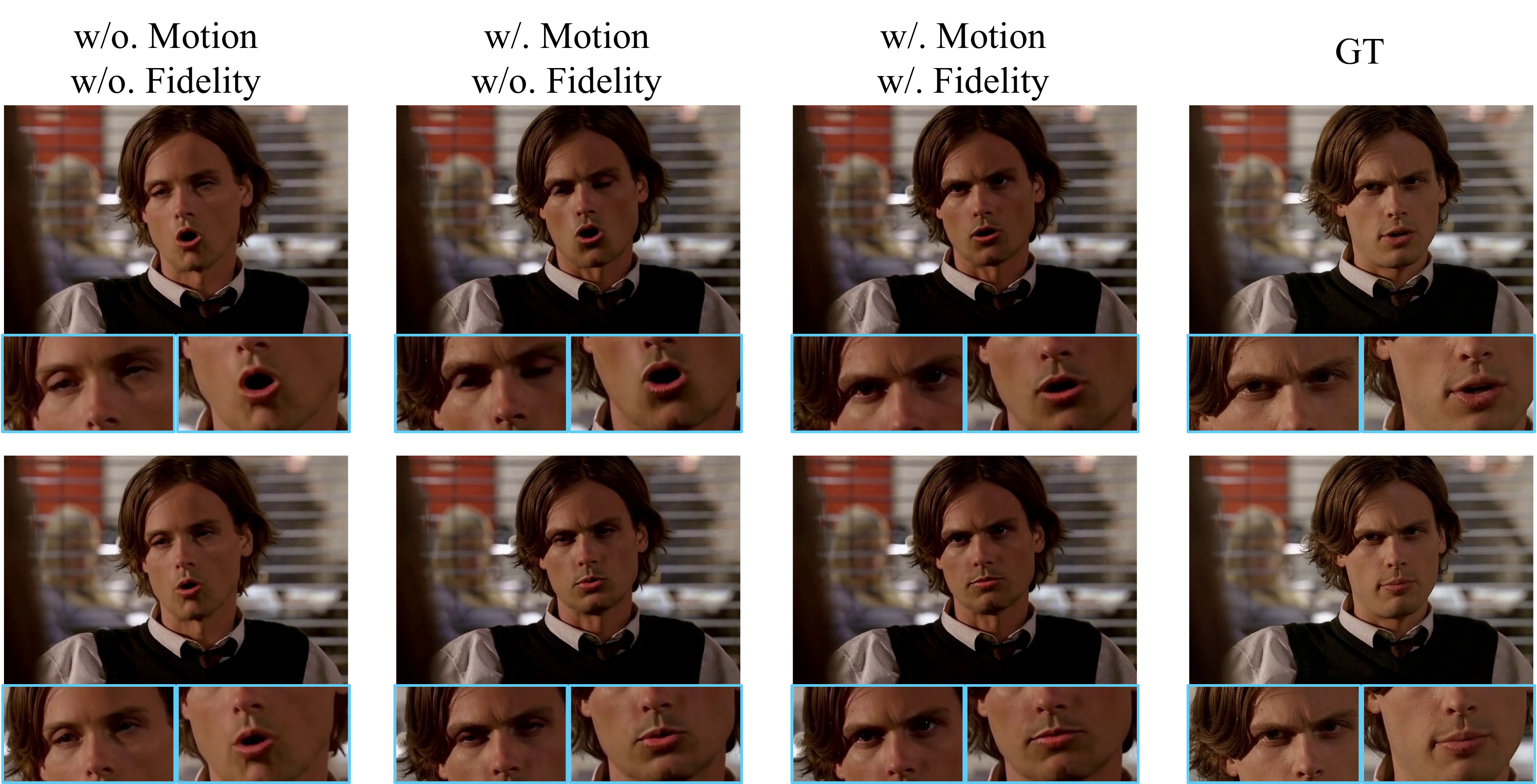}
    \caption{Ablation of direct preference optimization (with and without).}
    \label{fig:ablation_dpo}
\end{figure}

\textbf{Preference Metrics.}
As demonstrated in Table~\ref{tab:ablation_dpo} and Figure~\ref{fig:ablation_dpo}, integrating motion alignment preferences significantly improves synchronization metrics (Sync-C/Sync-D) compared to baseline models. 
Combining both motion alignment and portrait fidelity preferences achieves the optimal balance across synchronization (Sync-C: 5.689, Sync-D: 7.853) and expression quality (E-FID: 18.998), enhancing realism in generated animations while maintaining motion coherence.

\begin{table}[t!]
    \centering
    \begin{tabular}{cc|c|c|c|c|c}
        \toprule
        \textbf{Motion} & 
        \textbf{Fidelity} & 
        \textbf{Sync-C$\uparrow$} & 
        \textbf{Sync-D$\downarrow$} & 
        \textbf{E-FID$\downarrow$} & 
        \textbf{FID$\downarrow$} & 
        \textbf{FVD$\downarrow$} \\
        \midrule
        & &5.326 &8.391 &21.065 &59.007 &541.201 \\
        \checkmark & & 5.651 & 7.873 & 19.526 & \textbf{55.220} & \textbf{518.380} \\
        \checkmark & \checkmark & \textbf{5.689} & \textbf{7.853} & \textbf{18.998} & 58.815 & 538.396 \\
        \bottomrule
    \end{tabular}
    \vspace{2mm}
    \caption{Ablation of human preference metrics in direct preference optimization.
    ``Motion'' refers to the metric of human motion video alignment, and ``Fidelity'' refers to portrait fidelity.}
    \label{tab:ablation_dpo}
\end{table}

\begin{table}[t!]
    \centering
    \begin{tabular}{c|c|c|c|c|c}
    \toprule
        \ &
          \textbf{Sync-C$\uparrow$} &
          \textbf{Sync-D$\downarrow$} &
          \textbf{E-FID$\downarrow$} &
          \textbf{FID$\downarrow$} &
          \textbf{FVD$\downarrow$} \\
          \midrule
        better vs. worse &5.341 &8.197 &20.199 &58.947 &541.684 \\
        best vs. worse &5.506 &8.086 &19.990 &61.371 &\textbf{530.853} \\
        better vs. worst &5.401 &8.214 &19.944 &58.959 &588.209 \\
        \midrule
        best vs. worst &\textbf{5.689} &\textbf{7.853} &\textbf{18.998} &\textbf{58.815} &538.396 \\
        \bottomrule
    \end{tabular}
    \vspace{2mm}
    \caption{Ablation study of pairwise preference construction strategies on Celeb-V dataset. The ``best-vs-worst'' strategy, which pairs the highest- and lowest-quality samples, achieves superior synchronization (Sync-C/D) and expression fidelity (E-FID) compared to alternative pair selection methods.}
    \label{tab:ablation_pairwise}
\end{table}

\textbf{Pairwise Preference Construction Strategy.}
As show in Table~\ref{tab:ablation_pairwise}, we analyze these different preference data pair construction strategies. 
The ``better-worse" approach generates all possible pairs $(v_i, v_j)$ where $r_i > s_j$. The ``best-vs-worse" strategy adopts a top-down perspective, pairing the highest-scoring video with each lower-quality video. Conversely, the ``better-vs-worst" method employs a bottom-up framework by contrasting the lowest-quality sample against all superior videos. The ``best-vs-worst" strategy, used by us, pairs only the highest and lowest-scoring video. The experiments show that this strategy, which
generates only the most distinctive pair, yields the best performance.

\begin{table}[t!]
    \centering
        \begin{tabular}{c|c|c|c|c|c}
        \toprule
            \ &
              \textbf{Sync-C$\uparrow$} &
              \textbf{Sync-D$\downarrow$} &
              \textbf{E-FID$\downarrow$} &
              \textbf{FID$\downarrow$} &
              \textbf{FVD$\downarrow$} \\
              \midrule
            audio/4 &2.769 &10.852 &20.263 &\textbf{56.149} &538.821 \\
            audio/2 &4.244 &8.849 &19.464 &56.382 &557.557 \\
            \midrule
            Ours &\textbf{5.689}  &\textbf{7.853} &\textbf{18.998} & 58.815 &\textbf{538.396}  \\ 
            \bottomrule
        \end{tabular}
        \vspace{2mm}
        \caption{Ablation study of temporal motion modulation strategies for audio conditioning. 
        ``audio/4'' denotes temporal compression by 4× without channel expansion, directly aligning audio features with compressed latents. 
        ``audio/2'' applies 4× temporal compression followed by 2× channel expansion. 
        Our method employs 4× temporal compression with proportional 4× channel expansion, redistributing features to preserve temporal fidelity.} 
        \label{tab:ablation_audio_cond}
\end{table}

\begin{table}[t!]
    \centering
        \begin{tabular}{c|c|c|c|c|c|c}
        \toprule
        \ &
        \textbf{HKV$\uparrow$} &
        \textbf{HKC$\uparrow$} & 
        \textbf{PSNR$\uparrow$} &
        \textbf{SSIM$\uparrow$} &
        \textbf{FID$\downarrow$} &
        \textbf{FVD$\downarrow$} \\
        \midrule
        skeleton/4 &31.096&0.850 &20.752&0.789 &48.079&438.024 \\
        skeleton/2&33.567&0.873 &21.317&0.802&47.112&410.573 \\
        \midrule
        Ours & \textbf{35.736} &\textbf{0.886} & \textbf{22.221}&\textbf{0.808} & \textbf{44.422}& \textbf{381.752}\\ 
        \bottomrule
        \end{tabular}
        \vspace{2mm}
        \caption{Ablation study of temporal motion modulation strategies for skeleton conditioning. 
        ``skeleton/4'' denotes temporal compression by 4× without channel expansion, directly aligning skeleton features with compressed latents. 
        ``skeleton/2'' applies 4× temporal compression followed by 2× channel expansion. 
        Our method employs 4× temporal compression with proportional 4× channel expansion, redistributing features to preserve temporal fidelity.}
        \label{tab:ablation_unified_cond}
\end{table}

\begin{figure}
    \centering
    \includegraphics[width=1\linewidth]{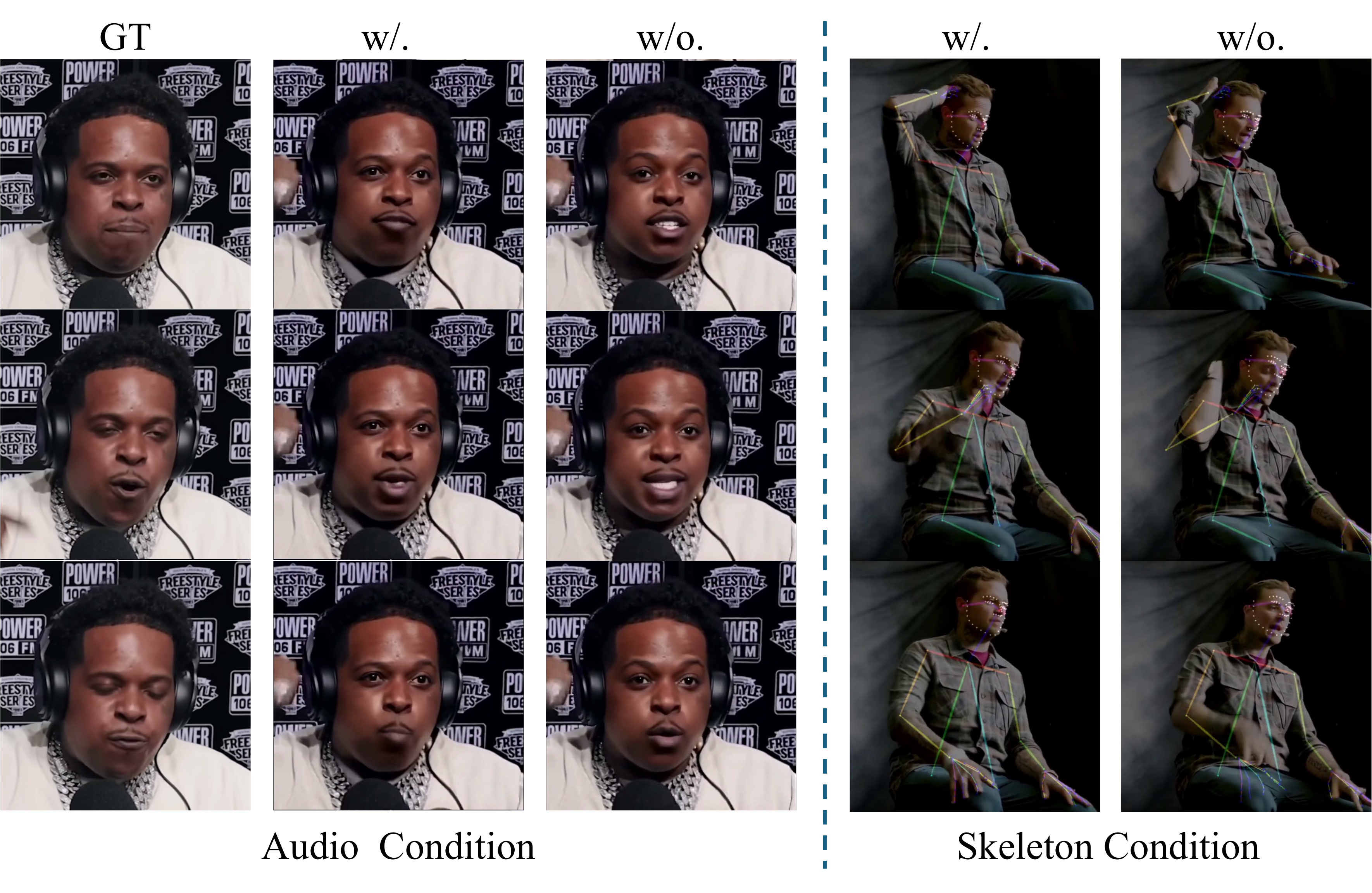}
    \vspace{2mm}
    \caption{Ablation of unified motion modulation (with and without).}
    \label{fig:unified_motion}
    \vspace{-4mm}
\end{figure}

\textbf{Temporal Motion Modulation for Audio Conditioning.}
To evaluate the efficacy of our temporal reshaping strategy for audio features, we conduct ablation studies comparing different downsampling approaches. 
As shown in Table~\ref{tab:ablation_audio_cond}, direct temporal compression (audio/4) without channel expansion significantly degrades lip synchronization (Sync-C: 2.769), while partial expansion (audio/2) improves Sync-C to 4.244. 
Our full approach—temporal compression with proportional 4× channel expansion—achieves optimal Sync-C (5.689) and E-FID (18.998) by preserving high-frequency speech dynamics through feature redistribution rather than subsampling.
Visualization results are shown in Figure~\ref{fig:unified_motion}.

\textbf{Temporal Motion Modulation for Skeleton Conditioning.}
Similar analysis on skeletal motion modulation (Table~\ref{tab:ablation_unified_cond}) reveals that direct temporal compression (skeleton/4) reduces hand motion diversity (HKV: 31.096). 
Partial expansion (skeleton/2) improves HKV to 33.567, while our full method achieves superior articulation quality (HKC: 0.886) and motion variety (HKV: 35.736) through proportional channel expansion. 
This demonstrates that redistributing temporal features into expanded channels better preserves anatomical details critical for expressive body animations.

\textbf{Comparison with Supervised Fine-tuning.}
In Table~\ref{tab:ablation_sft}, we compare direct preference optimization with supervised fine-tuning on Celeb-V. 
While SFT marginally improves Sync-C (5.387 vs. 5.326) and E-FID (21.432 vs. 21.065) over the baseline, DPO achieves superior gains (Sync-C: 5.689, E-FID: 18.998). 
This demonstrates that preference-driven training better aligns generated animations with human judgments than traditional loss-driven approaches, despite equivalent training data exposure.

\begin{table}[t!]
    \centering
    \begin{tabular}{c|c|c|c|c|c}
    \toprule
        \ &
          \textbf{Sync-C$\uparrow$} &
          \textbf{Sync-D$\downarrow$} &
          \textbf{E-FID$\downarrow$} &
          \textbf{FID$\downarrow$} &
          \textbf{FVD$\downarrow$} \\
          \midrule
        Baseline &5.326 &8.391 &21.065 &59.007 &541.201 \\
        SFT &5.387 &8.352 &21.432 &59.377 &540.443 \\
        Ours &\textbf{5.689} &\textbf{7.853} &\textbf{18.998} &\textbf{58.815} &\textbf{538.396} \\
        \bottomrule
    \end{tabular}
    \vspace{2mm}
    \caption{Comparison between direct preference optimization (Ours) and supervised fine-tuning (SFT) on Celeb-V dataset.}
    \label{tab:ablation_sft}
\end{table}

\begin{table}[!t]
    \centering
    \begin{tabular}{c|c|c|c|c} 
    \toprule
    \textbf{Method / Metric} & \textbf{Lip sync} & \textbf{Beat align} & \textbf{Portrait fid} & \textbf{Motion  div} \\ 
    \midrule
    FantasyTalking~\cite{wang2025fantasytalking} & 3.38 & 3.33 & 3.50 & 2.73 \\
    DreamTalk~\cite{ma2023dreamtalk} & 2.83 & 2.88 & 2.68 & 3.20 \\
    Hallo3~\cite{cui2024hallo3} & 3.90 & 3.73 & 3.93 & 3.80 \\
    Sonic~\cite{ji2024sonic} & 4.18 & 4.08 & 4.13 & 4.03 \\
    \midrule
    Ours & \textbf{4.60} & \textbf{4.48} & \textbf{4.43} & \textbf{4.23} \\ 
    \bottomrule
    \end{tabular}
    \vspace{2mm}
    \caption{User study comparison of lip synchronization, beat alignment, portrait fidelity and motion diversity on EMTD dataset}
    \label{tab::user-study}
\end{table}

\subsection{User Studies.}  
We conducted a user study evaluating 200 generated samples from the EMTD dataset across four critical dimensions: lip synchronization, beat alignment, portrait fidelity, and motion diversity. 
As shown in Table~\ref{tab::user-study}, our method achieves superior ratings compared to state-of-the-art baselines, with relative improvements of 10.1\% in lip synchronization and 5.0\% in motion diversity over the strongest baseline (Sonic). 
Participants rated samples on a 5-point scale, where our approach scored 4.43 for portrait fidelity – surpassing Hallo3 by 12.7\% – demonstrating its ability to synthesize identity-consistent animations with enhanced motion variety and visual coherence.

\subsection{Limitations and Future Work}
While our method demonstrates obvious improvements in lip and body synchronization and portrait fidelity, several limitations remain. 
First, the current framework requires paired audio-visual data with precise temporal alignment for optimal performance, which may constrain its applicability to unconstrained real-world scenarios. 
Second, the unified motion modulation mechanism, while effective, introduces additional computational overhead during training due to feature channel expansion. 
Third, our preference dataset currently focuses on upper-body animations, leaving full-body motion synthesis as an open challenge.

Future work will explore three key directions: 
(1) extending the preference optimization framework to handle weakly aligned or unlabeled data through self-supervised learning, 
(2) developing more efficient temporal modulation strategies to reduce computational costs, 
and (3) expanding the motion condition framework to support full-body dynamics while maintaining fine-grained control over facial expressions and hand gestures. 
These advancements could further bridge the gap between synthetic animations and natural human movements.

%% file: section/conclusion.tex
\section{Conclusion}
We present a novel framework for human preference-aligned portrait animation that integrates direct preference optimization with unified temporal motion modulation. 
By curating the first DPO dataset targeting motion-video synchronization and expression naturalness, our method significantly enhances lip-sync accuracy and facial expressiveness while maintaining high-fidelity rendering. 
The proposed temporal modulation mechanism effectively resolves motion granularity degradation in DiT architectures through feature redistribution, enabling precise synchronization of rapid articulatory gestures. 
While currently focused on upper-body synthesis, our work establishes a foundation for extending preference-aware optimization to full-body portrait animation. 